\RequirePackage{fix-cm}
\documentclass[natbib,twocolumn]{svjour3}          
\usepackage[latin1]{inputenc}
\smartqed  
\usepackage{graphicx}
\usepackage[british]{babel}
\usepackage{booktabs} 
\usepackage{array} 
\usepackage{paralist} 
\usepackage{verbatim} 
\usepackage{enumitem}
\usepackage{subfig}

\usepackage{multirow} 
\usepackage{arydshln} 
\usepackage{amsfonts} 
\usepackage{amssymb,amsmath} 
\DeclareMathOperator*{\argmax}{argmax}
\usepackage{mathtools}

\usepackage{pifont}
\newcommand{\xmark}{\ding{55}}%

\usepackage{colortbl}
\usepackage{xcolor}
\definecolor{lightgray}{gray}{10}
\usepackage{natbib}
\usepackage{lipsum} 

\begin{document}\sloppy

\title{Hypothesis-based Belief Planning for Dexterous Grasping\thanks{We gratefully acknowledge support of FP7 grant IST-600918, PacMan.}
}


\author{Claudio Zito \and
         Valerio Ortenzi \and Maxime Adjigble \and Marek Kopicki \and Rustam Stolkin \and Jeremy L. Wyatt
}


\institute{Zito, Adjigble, Kopicki, Stolkin and Wyatt\at
              University of Birmingham, UK \\
              Ortenzi\at
              Centre of Excellence, Queensland University of Technology, AU\\
              \email{\{claudio.zito.81\}@gmail.com}}

\date{Received: date / Accepted: date}

\maketitle

\begin{abstract}

Belief space planning is a viable alternative to formalise partially observable control problems and, in the recent years, its application to robot manipulation problems has grown. However, this planning approach was tried successfully only on simplified control problems.  
In this paper, we apply belief space planning to the problem of planning dexterous reach-to-grasp trajectories under object pose uncertainty.
In our framework, the robot perceives the object to be grasped on-the-fly as a point cloud and compute a full 6D, non-Gaussian distribution over the object's pose (our belief space). The system has no limitations on the geometry of the object, i.e., non-convex objects can be represented, nor assumes that the point cloud is a complete representation of the object. A plan in the belief space is then created to reach and grasp the object, such that the information value of expected contacts along the trajectory is maximised to compensate for the pose uncertainty. If an unexpected contact occurs when performing the action, such information is used to refine the pose distribution and triggers a re-planning. Experimental results show that our planner (IR3ne) improves grasp reliability and compensates for the pose uncertainty such that it doubles the proportion of grasps that succeed on a first attempt.

\keywords{Dexterous Grasping \and Information Gain \and Belief Space Planning \and Reach to Grasp Planning}
\end{abstract}

\begin{figure}[!t]
\centerline{
\includegraphics[width=0.98\columnwidth]{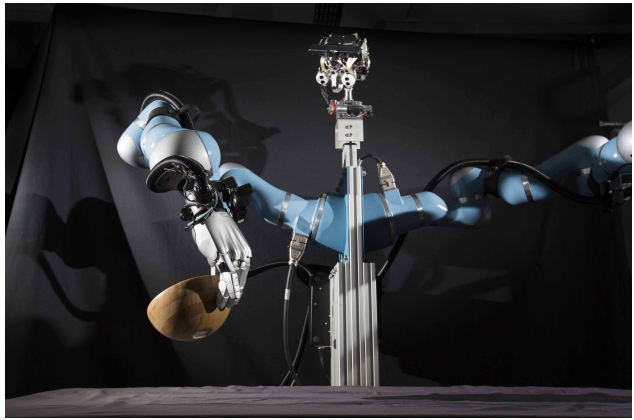}
}
\caption[Boris]{Boris: half-humanoid robot platform developed at the University of Birmingham.}
\label{fig:boris}
\end{figure}

\section{Introduction}
\label{sec:intro}


Imagine that you are reaching into the fridge to grasp an object you can only partially see. Rather than relying solely on vision, you must use touch in order to localise it and securely grasp it. 
However, humans would not poke the object to localise it first and then grasp it. We compensate for the uncertainty by approaching the object in a way such that if a contact occurs it will generate enough information about where the object is and the object will be grasped with a minimum adaptation of the initial trajectory.

Previous work attempted to couple the uncertainty reduction and grasp execution in the framework of partially observable Markov decision processes (POMDPs). Although methods are advancing for continuous state \citep{porta2006point,bai2010monte,brooks2006parametric} and continuous action spaces \citep{porta2006point,murphy2000survey}, no POMDP planners yet scale to the high dimensional belief state and action spaces required for robot grasping. This is especially true for manipulators with great dexterity as they have a high number of degrees of freedom. Instead, actual robot implementations of the POMDP approach to active tactile grasping separate exploratory and grasping actions and plan how to sequence them \citep{hsiao2011robust}, typically by relying on a user-defined threshold to know when the belief is accurate enough to switch from gathering information to execution of a pre-computed grasp action. This approach fails to exploit the fact that, in tactile grasping, hand movements can both move towards the grasp configuration, and reduce uncertainty. They are most naturally performed concurrently, rather than sequentially. Furthermore, these approaches typically rely on constraining the belief space to Gaussian distributions. Extensions to non-parametric representations of belief~\citep{bib:nikandrova_2014}, typically result in intractable planning problems due to the high dimensionality of non-Gaussian parametrisation.

This work presents a formulation of dexterous manipulation that aims to exploit concurrency in exploratory and grasping actions for reach-to-grasp hand movements. The properties of the approach are that it:
\begin{itemize}
\item tracks high-dimensional belief states defined over a 6D, non-Gaussian pose uncertainty;
\item efficiently plans in a fixed-dimensional space;
\item simultaneously gathers information while grasping, i.e., there is no need to switch between gathering information and grasping since the action space is the space of dexterous reach-to-grasp trajectories;
\item does not require a user-supplied mesh model of the object or a pre-computed grasp associated with the object;
\item copes also with non-convex objects, i.e., there are no limitations to the shape of the objects that it can successfully grasp.
\end{itemize} 

We build our approach by combining the idea of hypothesis-based planning (HBP), initially proposed in \citep{bib:platt_isrr_2011}, and our one-shot learning algorithm for dexterous grasping of novel objects \citep{bib:kopicki_2015}. The hypothesis-based planner works as follows. Instead of planning directly in a high dimensional belief space, our plan is constructed on a fixed-dimensional, sampled representation of belief. In other words, the belief space is projected onto a set of competing hypothesis in the underlying state space. However, our implementation of the HBP algorithm extends the work in \citep{bib:platt_isrr_2011} in several directions. First, Platt's formulation of the hypothesis-based planner is defined on a set of actions (i.e., movement constrained in the horizontal plane) that differs from the actual grasp (i.e. a pinch grasp with two paddles). In contrast, we formulate the problem on the same action space for each stage (i.e., dexterous reach-to-grasp trajectories). As a result, we do not require a user-supplied threshold over the current belief to estimate when to interrupt the information gathering phase and execute a pre-defined grasp. Another difference is that the observational model used in~\citep{bib:platt_isrr_2011} relies on contactless sensors (i.e. laser sensors), while we maximise tactile observations for a dexterous robotic hand; and, finally, we do not make any assumptions about the model of the object to be grasped, in contrast to the original work that assumes a convex object (i.e. a box) aligned in front of the robot (see Sec.~\ref{sec:grasp_planning} for further details).
On top of our hypothesis-based planner, our grasping algorithm enables us to learn a set of grasp contacts on a point-cloud object model (PCOM), directly obtainable from a depth camera, and to generalise to objects of novel shape. Therefore we do not require a mesh model of the object, and we can also generate grasps on target objects with incomplete point clouds. Hence our algorithm is exceptionally flexible in planning dexterous reach-to-grasp trajectories for novel objects.

\begin{figure}[!t]
\centerline{
     \includegraphics[width=0.95\columnwidth]{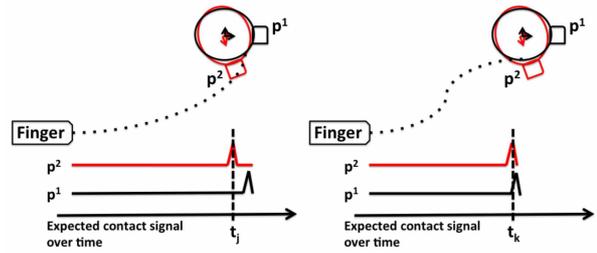}\label{fig:obs_model_1}\label{fig:obs_model_2}
}
\caption[Observational model]{The figures show the observational model for tactile information. The poses $p_1$ and $p_2$ represent two hypothesised configurations of a mug to be grasped. The dotted lines show two possible trajectories for the finger to reach and touch the mug. Hypothesis $p_1$ represents the expected mean pose for the mug. The left figure shows the expected contact signal for both hypotheses along the trajectory. At time $t_j$ the planner expects to observe a contact if the object is in pose $p_2$ and no contact for pose $p_1$. In the right figure, the planner expects similar observations in both cases at time $t_k$. Thus the trajectory on the left is more likely to distinguish hypothesis $p_1$ versus $p_2$ than the trajectory on the right.}
\label{fig:obs_model}
\end{figure}

In order to link these two methods, hypothesis based planning and dexterous grasping of novel objects, we need to construct a representation of the belief space that will allow us to track pose uncertainty for a PCOM, in 6D, and cope with the non-Gaussian posterior.  We do so by employing a non-parametric representation of the belief state defined as a kernel density estimator. Each kernel is a weighted pose of the target object inferred from visual data collected on the fly (see Sec.~\ref{sec:belief_state_estimation}).

Our experimental results show that our planner, IR3ne, is more reliable than open-loop grasping from vision. We further show that IR3ne improves over simple tactile re-planning in three ways: i) it doubles the proportion of times in which the robot reaches and grasps the object at the first attempt, ii) if an unexpected contact is generated along the trajectory, the contact provides more information about the object's pose, and thus iii) it reduces the number of replanning steps before converging to a grasp. Experiments in simulation and on a real robot (Fig.~\ref{fig:boris}) confirm these attractive properties.

We continue with related work (Sec.~\ref{sec:background}), then describe the planning problem formulation (Sec.~\ref{sec:problem}). We describe all aspects of the framework (Sec.~\ref{sec:implementation}), and report and discuss the experimental results (Sec.~\ref{sec:experiments}). We finish with concluding remarks (Sec.~\ref{sec:conclusion}).

\begin{table}
\centering
\caption{IR3ne vs i) the hypothesis-based planner (HBP) proposed in \citep{bib:platt_isrr_2011}, ii)  the informative sensor-based grasp planner (ISBP) proposed in \citep{bib:nikandrova_2014},  and iii) the POMDP-based approach proposed in \citep{hsiao2011robust}}
\label{tab:comparison}
\scriptsize
\begin{tabular}{lcccccccccc}
  & 
 \rotatebox{90}{\scriptsize Tracking high-dimensional belief} & 
\rotatebox{90}{\scriptsize 6D non-Gaussian pose uncertainty} & 
\rotatebox{90}{\scriptsize Planning in a fixed-dimensional space} & 
 \rotatebox{90}{\scriptsize Actively gathering tactile information} & 
 \rotatebox{90}{\scriptsize Simultaneously IG \& grasping} & 
 \rotatebox{90}{\scriptsize Switching IG \& grasping}& 
\rotatebox{90}{\scriptsize Planning for dexterous grasp} & 
\rotatebox{90}{\scriptsize Coping with non-convex objects}  & 
\rotatebox{90}{\scriptsize No need for user-supplied grasp} & 
 \rotatebox{90}{\scriptsize Recovering from incorrect plans}\\
 \hline
 \rowcolor{gray!10}
HBP & \checkmark & \xmark &\checkmark & \xmark & \checkmark & \xmark & \xmark & \xmark & \xmark & \checkmark\\
ISBP & \checkmark & \xmark & \xmark & \xmark & \xmark & \checkmark  & \xmark & \xmark & \xmark & \checkmark \\
 \rowcolor{gray!10}
 POMDP & \checkmark & \xmark & \xmark & \checkmark & \xmark & \xmark & \xmark &  \xmark & \xmark & \checkmark\\
IR3ne & \checkmark & \checkmark & \checkmark & \checkmark & \checkmark & \checkmark & \checkmark & \checkmark & \checkmark & \checkmark\\\hline
\end{tabular}
\end{table}

\section{Related Work}
\label{sec:background}


The problem of robot grasping under object pose uncertainty can be decomposed into: state estimation, grasp synthesis, reach to grasp planning, and control. A typical approach is to represent the belief state using prior distributions, select a grasp robust w.r.t. uncertainty and finally use tactile feedback to adjust the grasping trajectory, see e.g.~\citep{bib:nikandrova_2014}. The reach-to-grasp trajectory is generally computed using sampling-based techniques which minimise the trajectory cost. Less work has explored the more complex problem of reasoning about uncertainty while planning this dexterous reach-to-grasp trajectory. 

Table~\ref{tab:comparison} summarises the main differences between our approach and the closest related work at glance. 

\subsection{State estimation}\label{sec:state_estimation}

One class of approaches to state estimation uses a \emph{maximum likelihood estimate} (MLE), e.g.~\citep{bib:kopicki_2014}. Given an object model, a typical robust global estimator samples subsets of points and computes hypothesised poses based on feature correspondences between the data and the model~\citep{bib:hillenbrand_2008, bib:tuzel_2005, bib:uli_cviu_2011, bib:bracci_2017}. Nonetheless, even for full shape matching, small errors in the pose estimate may lead to critical failures while attempting to grasp, i.e., unexpected contacts may damage the object or the hand itself. Another class of approaches maintains a \emph{belief state}, which is a probability distribution over the underlying states. Maintaining a density over the pose of the object yields Bayesian strategies that are capable of maximising the probability of grasp success given the current belief state, as in~\citep{bib:hsiao_icraw_2011}. However, belief space planning can be computationally expensive, especially for high-dimensional, complex density functions. A popular choice is to constrain the belief space to Gaussian density functions. 

Since many robotic problems have multi-modal uncertainty, there are benefits in using a non-parametric representation of the belief space, such as a \emph{particle filter} (PF). There are many examples of PFs used for state estimation in manipulation problems~\citep{bib:petrovskaya_trans_2011},~\citep{bib:nikandrova_2014} and~\citep{bib:platt_csail_2011}. Most of these methods sample an initial particle set from an user-defined distribution attached to the \emph{maximum a posteriori} estimate (MAP) obtained from vision, and often the uncertainty is limited to 2D. 

This paper employs a pair-fitting model similar to that described in~\citep{bib:uli_cviu_2011} to estimate the possible poses from RGB-D data, where each hypothesis is an MLE. A detailed explanation is given in Sec.~\ref{sec:belief_state_estimation}.

\subsection{Grasp Synthesis}\label{sec:grasp_synthesis}

There are analytic and empirical approaches to grasp synthesis \citep{bib:sahbani_2012}. Analytic approaches are typically associated with an optimisation problem, and thus the computational effort grows with the size of the grasp solution space, which in turn grows with the number of fingers and contact points. On the other hand, data driven approaches learn a mapping from an object description to the grasp \citep{bib:saxena_2008, bib:detry_2010, bib:kopicki_2014, bib:gu_2017}. Learning this mapping from incomplete or erroneous data remains challenging. In~\citep{bib:detry_2010}, the authors address the problem of associating a grasp with a partially perceived object. 

In our previous work~\citep{bib:kopicki_2014, bib:kopicki_2015}, an efficient method is presented to learn a dexterous grasp type (e.g. pinch, rim) from a single example. The method can generalise within and across object categories, and use full or partial shape information. In the original work in~\citep{bib:kopicki_2014, bib:kopicki_2015}, each contact model is associated with an (open-loop) approaching trajectory demonstrated during the learning phase. In this paper, we employ the same contact model described in~\citep{bib:kopicki_2015} but we do not use a purely open loop grasp method. Instead we replace it with the trajectory computed by our planner, and recomputed in the case of feedback from tactile contact. 


\subsection{Grasp Planning}\label{sec:grasp_planning}

Some algorithms pose the reach to grasp problem as a belief-state planning problem. Typical implementations rely on constraining the belief space to a Gaussian. Extensions to non-parametric representations of belief~\citep{bib:nikandrova_2014}, typically result in intractable planning problems due to the high dimensionality of the non-Gaussian parametrisation. In contrast, Platt et al.~\citep{bib:platt_csail_2011}, ~\citep{bib:platt_icra_2012}, enabled efficient planning under state uncertainty. The key was to plan an action sequence which will generate observations that distinguish between competing hypotheses. 
However, this was formulated and tested for bi-manually grasping a cardboard box, using a laser range finder on one arm as the sensor. All movements of the arms were constrained to lie in the horizontal plane, as was the object pose uncertainty. With this formulation, a hypothesis based plan was created to gather information, while aligning the robot's arm equipped with the laser in front of the box's edge. The system required a user-supplied model of the object and relied on a user-defined threshold to know when the belief is accurate enough to switch from gathering information to executing a pre-computed grasp action. Another difference with our work is that Platt's formulation of the hypothesis-based planner is defined on a set of actions (i.e., left-right movement of the left arm of the robot) that are distinct from the actual grasping action (i.e., a pinch grasp with two paddles).
In contrast, we formulate the grasping and information gathering problems on the same action space (i.e., dexterous reach-to-grasp trajectories) and our system can dextrously grasp non-convex objects with non-Gaussian 6D uncertainty.

There is also evidence that humans compensate for uncertainty due to noisy motor commands and imperfect sensory information during the execution of reach-to-grasp movements~\citep{bib:kording_2004}. The authors in~\citep{bib:christopoulos} induced  pose uncertainty on the object to be grasped in order to investigate the compensation strategies adopted by humans. The authors evaluated the benefits of uncertainty compensation, based on the hypothesis that participants prefer to generate force-closure grasps at first contact, and therefore the participants tended to modify their approach along the direction of maximum uncertainty and increase their peak grip-width. The experimental results in our paper show that our information reward algorithm, IR3ne, yields a similar behaviour.

%

\section{Problem Formulation}
\label{sec:problem}

\begin{figure}[t]
\centerline{
\includegraphics[width=0.98\columnwidth]{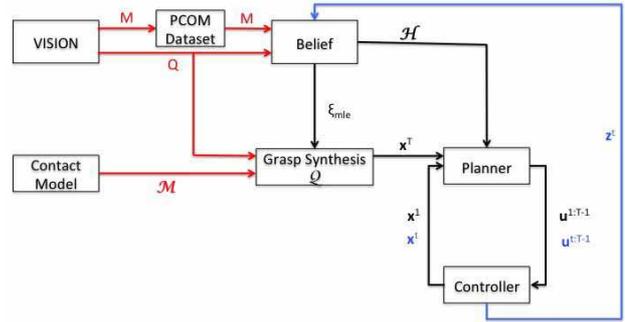}
}
\caption{The algorithm at glance. Red lines shows the input to the system only at the beginning of the execution. From vision two point clouds are obtained: i) the PCOM $M$, which is stored into the PCOM database, and ii) the object's point cloud. A contact model, $\mathcal{M}$, computed off-line with our one-shot learning algorithm. Black lines define the flow of the algorithm. First, a belief state is estimated by aligning $M$ onto $Q$. The maximum likelihood estimate, $\xi_{mle}$ is used to compute the target grasp $\mathbf{x}^T$ on $Q$. Our planner computes a reach-to-grasp trajectory from the current state of the robot, $\mathbf{x}^1$, to $\mathbf{x}^T$, which compensate for the uncertainty by reasoning on the set of hypotheses, $\mathcal{H}$. The controller executes the trajectory, $\mathbf{u}^{1:T-1}$. If a contact occurs at time $t$, the controller stops the robot and the replanning phase is triggered. The blue line shows the contact signal, $\mathbf{z}^t$ which is integrated in the posterior. The algorithm computes a new target grasp and set of hypotheses and the planner computes a new trajectory from state $\mathbf{x}^t$.}
\label{fig:algorithm}
\end{figure}

In this section, we formulate our information-based planner for dexterous robot grasping under pose uncertainty. Figure~\ref{fig:algorithm} shows our algorithm at glance.


Let us consider a discrete-time system with continuous, non-linear, deterministic dynamics, 
\begin{equation}\label{eq:f}
\mathbf{x}^{t+1}=f(\mathbf{x}^t,\mathbf{u}^t) 
\end{equation}
where $\mathbf{x}^t\in\mathbb{R}^n$ is a configuration state of the robot and $\mathbf{u}^t\in\mathbb{R}^l$ is an action vector, both parameterised over time $t\in\{1,2,...\}$. Our goal is to find a trajectory, $\mathbf{u}^{1:T-1}$, the moves our dexterous robot manipulator from its current pose, $\mathbf{x}^1$, to a target grasp, $\mathbf{x}^T$ while compensating for the object pose uncertainty.
Thus the three following sub-problems are required to be solved:
\begin{enumerate}
\item How to estimate the object pose uncertainty.
\item How to compute the target grasp, $\mathbf{x}^T$, on an unknown object with pose uncertainty.
\item How to compensate for the uncertainty along the reach-to-grasp trajectory, $\mathbf{u}^{1:T-1}$.
\end{enumerate} 

\begin{table}
\centering
\caption{List of symbols}
\label{tab:symbols}
\scriptsize
\begin{tabular}{| l | l |}
\hline
$b$ & Density distribution over the object's pose (belief state) \\ \hline
$f(\cdot)$ & Non-linear system \\ \hline
$F_t(\cdot)$ & Robot's pose at time t \\ \hline
$g(\cdot)$ & Observation function \\ \hline
$g_t(\cdot)$ & Expected observation at time $t$ \\ \hline
$\mathcal{H}$ & Set of hypotheses subsampled from $b$ \\ \hline
$k$ & Number of particles in $\mathcal{H}$ \\ \hline
$K$ & Number of particles to approximate $b$ \\ \hline
$\mathcal{J}(\cdot)$ & Cost function for planning \\ \hline
$M$ & Model point cloud (PCOM) \\ \hline
$\mathcal{M}$ & Contact model for a grasp \\ \hline
$^OM$ & Reference frame for $M$ in world coordinates\\ \hline
$N$ & Number of particles in $b$ \\ \hline
$\mathcal{S}_1$ & Initial set of object's poses \\ \hline
$t$ & Time step in $[1,\cdots,T]$\\ \hline
$T$ & Final time step \\ \hline
$p$ & Elements of $\mathcal{H}$ \\ \hline
$p_1$ & MLE hypotheses in  $\mathcal{H}$ \\ \hline
$p_{2:k}$ & Hypotheses in  $\mathcal{H}$ sampled from $b$ \\ \hline
$Q$ & Query point cloud \\ \hline
$\mathbb{Q}$ & Block diagonal of measurement noise covariance matrix \\ \hline
$\mathcal{Q}$ & Contact query for a grasp \\ \hline
$^OQ$ & Reference frame for $Q$ in world coordinates\\ \hline
$\mathbf{u}$ & Action vector \\ \hline
$\mathbf{u}^{1:T-1}$ & Robot's trajectory \\ \hline
$\mathbf{x}$ & Robot's pose in joint space \\ \hline
$\mathbf{x}^T$ & Ending of the reach-to-grasp trajectory (target grasp) \\ \hline
$y$ & Pose in SE(3) in world coordinate \\ \hline
$\mathbf{z}$ & Tactile observation vector\\ \hline
$xi$ & Pose in SE(3) relative to $^OM$ \\ \hline
\end{tabular}
\end{table}

\subsection{Belief state estimation}\label{sec:belief_state_estimation}

To model multi-modal uncertainty in the object pose, we employ a non-parametric representation (PF) of the belief state, $b$, defined as a density function in SE(3). Let us denote by $SE(3)$ the Special Euclidean group of 3D poses (a 3D position and 3D orientation). An element $a\in SE(3)$ is a homogeneous transformation. Previous works typically initialised the belief by using an MLE of the object pose (i.e., from vision) and then manually generating a set of particles by sampling from a Gaussian distribution attached to the MLE. In contrast, our approach uses a sample-based model-fitting procedure similar to the one presented in~\citet{bib:uli_cviu_2011} to generate a set of MLEs, which is used as an initial belief state for the system. 
Thus, our belief state is defined non-parametrically by a set of $K$ 3D poses (or particles) $\mathcal{S}=\{(y_j, w_j)|y_j\in SE(3), w_j\in\mathbb{R}^+\}_{j=1}^K$.  
The probability density in a region is determined by the local density of the particles. The underlying pdf is approximated through a KDE~\citep{bib:silverman_1986}, by assigning a kernel function $\mathcal{K}$ to each particle supporting the density, as 
\begin{equation}\label{eq:density}
b(y)=\sum_j{w_j\mathcal{K}(y|y_j,\sigma)}
\end{equation} 
where $b(y)$ is the density value associated with the pose $y$, $\sigma$ is a diagonal matrix that represents the kernel bandwidth and $w_j$ is a normalised weight associated to $y_j$ such that $\sum_j{w_j}=1$. An initial set of poses, $\mathcal{S}_1$, is constructed by employing a pair-fitting model as explained below. However, it is assumed that a model of the object as a point cloud (PCOM) is available to the system prior to execution, see Sec.~\ref{sec:pcom}. 


The model-fitting procedure we employ to create an initial belief state is a surflet-pair fitting procedure~\citep{bib:uli_cviu_2011}. This algorithm is well-known in the computer vision community and works as follows. Let us consider the case in which a \emph{model point-cloud}, $M$, is available to the system, or collected on-the-fly, and a new \emph{query point cloud}, $Q$, is acquired. Let us also assume that each point and normal of $M$ is described w.r.t. an arbitrary reference frame, or pose, $^OM\in SE(3)$ in world coordinates. The goal of the algorithm is to search for a rigid body transformation $\xi_{mle}\in SE(3)$, described w.r.t. $^OM$, that best aligns $M$ with $Q$. This is achieved by sampling a set of surflet features for each point cloud (i.e. pairs of points with their relative normals) and aligning the features on $Q$ with the most similar in $M$. The algorithm also produces a score value to describe the \emph{goodness} of the alignment. We repeat this procedure $N$ times to construct our initial belief state, see Eq.({\ref{eq:density}), as a set of particles 
$$
\mathcal{S}_1=\{(y_j,w_j)|y_j=\xi^j_{mle}\times{^OM}\in SE(3)\}_{j=1}^N
$$
where $\xi^j_{mle}$ is the $j^{th}$ estimate created by the fitting procedure, $w_j$ is the associated score, $\times$ is a product of two homogeneous transformations and $y_j$ is a candidate pose of the object in our world coordinate system.

\subsection{Acquisition of a PCOM}\label{sec:pcom}

Our initial belief state estimation is created from visual data and we employ a sample-based model-fitting procedure, Sec.~\ref{sec:belief_state_estimation}, which relies on a PCOM available to the system before execution. Nonetheless, the acquisition of a PCOM is done by scanning the workspace of the robot with a depth camera. For novel objects, when the associated PCOM is not already present in the dataset, the system can acquire the PCOM on the fly and use the same point cloud as the model $M$ and query $Q$. Although it may seem redundant to compute the pose density on the same point cloud since one would expect to obtain a good fit when comparing a point cloud with itself, the benefit of maintaining a density over the pose estimation overcomes the overhead computation, as shown in our results, Sec.~\ref{sec:experiments}.  

It is important to notice that, in the current implementation of the system, the PCOM dataset contains only the model of the object to be grasped. Future work will focus on constructing a proper dataset of point clouds from which the system could generate a density function that would also take into account the shape uncertainty. However, this is outside the scope of this paper that only focus on demonstrating the benefits of compensating for pose uncertainty in dexterous reach-to-grasp actions. 

\subsection{The target grasp}

In order to create a plan that compensate for uncertainty we need to associate a grasp to the maximum likelihood pose of the object to be grasped. From Sec.~\ref{sec:belief_state_estimation}, we expect the object's point cloud, $Q$, to be in pose $\xi_{mle}=\mathbb{E}[b]$.

The target grasp is computed by using our one-shot grasping algorithm. This algorithm enables us to learn a contact model, $\mathcal{M}(\mathbf{x}, O)$, for a type of grasp (i.e., pinch with support) that describes the distribution of the contacts between the robot's end effector, $\mathbf{x}$, and the object's surface, $O$, with pose $\xi_O$, from a demonstration. We can compute  $\mathcal{M}(\mathbf{x}, O)$ off-line and store it in the system. Given a novel point cloud, $Q$, with pose $\xi_{mle}$, a new density function, $\mathcal{Q}(\mathcal{M}(\mathbf{x}, O), Q)$ is obtained for the new object. A candidate grasp can be sampled from $\mathcal{Q}$, so that $\mathbf{x}^T\sim\mathcal{Q}$. For further details refer to \citep{bib:kopicki_2015}. Figure~\ref{fig:07:bowllarge_pinchsupp} shows a pinch with support grasp learned on a bowl and transferred to a jug.    

\begin{figure}[!t]
\centerline{
\includegraphics[width=.95\columnwidth]{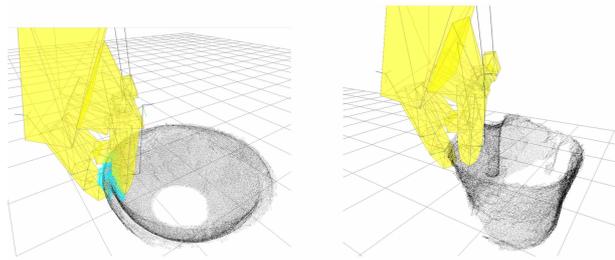}
}
\caption[Pinch with support grasp]{Pinch with support grasp learned on a bowl and transferred to a jug, using the method described in~\citep{bib:kopicki_2014}. The image on the left shows the learned contact model (blue points) for all the fingers involved in the grasp. The right image shows a possible grasp adaptation on a jug.}
\label{fig:07:bowllarge_pinchsupp}
\end{figure}

\subsection{Planning a trajectory to maximise information gain}

We know that in general the problem of planning in belief space is intractable~\citep{bib:littman}. Instead, let us consider a method to search for a sequence of actions, $\mathbf{u}^{1:T-1}=\{\mathbf{u}^1,\dots,\mathbf{u}^{T-1}\}$, that distinguish between observations that would occur if the object were in the expected pose, $\mathbb{E}[b]$, from those that would occur in other poses.\footnote{After the initial estimation of the belief state we do not rely anymore on vision. The only observations available to the system during the execution and re-planning are tactile observations.} See Fig.~\ref{fig:obs_model} for a graphical representation in 2D. A large number of particles, $K$, is suitable to track the belief state and converge to the true state of the system. Nevertheless, to create a plan, it is more convenient to sub-sample the hypotheses. Thus, we define a subset of $k$ hypotheses, 
\begin{equation}\label{eq:hypotheses}
\mathcal{H}=\{p_i\}_{i=1}^k
\end{equation}
such that $p_1=p_{mle}=\argmax[b]$ is the maximum likelihood estimate and $p_i\sim b$, with $i\in[2,k]$. 



The system in Eq.(\ref{eq:f}) also produces continuous, non-linear, observation dynamics,
\begin{equation}\label{eq:h}
\mathbf{z}_i^t=g(\mathbf{x}^t,p_i)
\end{equation}
where $\mathbf{z}_i^t\in\mathbb{R}^m$ is a vector that describes the expected observation (contacts), at time $t$, between the joint state of the robot, $\mathbf{x}$, and the object in pose $p_i$.
More generally, let $F_t(\mathbf{x},\mathbf{u}^{1:t-1})$ be the robot configuration at time $t$ if the system begins at state $\mathbf{x}$ and takes action $\mathbf{u}^{1:t-1}$. Therefore, the expected sequence of observations over a trajectory, $\mathbf{u}^{1:t-1}$, is:
\begin{multline}
g_t(\mathbf{x},\mathbf{u}^{1:t-1},p_i)=[g(F_2(\mathbf{x},\mathbf{u}^1),p_i)^\top, \\ g(F_3(\mathbf{x},\mathbf{u}^{1:2}),p_i)^\top, \ldots,g(F_t(\mathbf{x},\mathbf{u}^{1:t-1}),p_i)^\top]^\top
\end{multline}
a column vector which describes the expected contacts at any time step of the trajectory $\mathbf{u}^{1:t-1}$.
We then need to search for a sequence of actions which maximise the difference between observations that are expected to happen in the sampled states, $p_{2:k}$, when the system is actually in the most likely hypothesis, $p_1$. In other words, we want to find a sequence of actions, $\mathbf{u}^{1:T-1}$, that minimises
\begin{multline}\label{eq:J_cost}
\mathcal{J}(\mathbf{x},\mathbf{u}^{1:T-1},p_{1:k})= \\ \sum_{i=2}^kN(g_T(\mathbf{x},\mathbf{u}^{1:T-1},p_i)|g_T(\mathbf{x},\mathbf{u}^{1:T-1},p_1),\mathbb{Q})
\end{multline}
where $N(\cdot|\mu,\Sigma)$ denotes the Gaussian distribution with mean $\mu$ and covariance $\Sigma$ and $\mathbb{Q}$ is the block diagonal of the measurement noise covariance matrix. Rather than optimising~(\ref{eq:J_cost}) we follow the suggested simplifications described in~\citep{bib:platt_csail_2011} by dropping the normalisation factor in the Gaussian and optimising the exponential factor only. Let us define for any $i\in[2,k]$
\begin{multline}
\Phi(\mathbf{x},\mathbf{u}^{1:T-1},p_i)= \\ ||g_T(\mathbf{x},\mathbf{u}^{1:T-1},p_i)-g_T(\mathbf{x},\mathbf{u}^{1:T-1},p_1)||_{\mathbb{Q}}^2
\end{multline}
then the modified cost function is
\begin{equation}\label{eq:modifiedcost1}
\mathcal{J}(\mathbf{x},\mathbf{u}^{1:T-1},p_{1:k})=\frac{1}{k}\sum_{i=2}^k{e^{-\Phi(\mathbf{x},\mathbf{u}^{1:T-1},p_i)}}
\end{equation}
it is worth noting that, when there is a significant difference between the two sequences of expected observations, $g_T(\mathbf{x},\mathbf{u}^{1:T-1},p_i)$ and $g_T(\mathbf{x},\mathbf{u}^{1:T-1},p_1)$, the function $\Phi(\cdot)$ is large and therefore $\mathcal{J}(\cdot)$ is small. On the other hand, if the two sequences of expected observations are very similar to one another, their distance tends to 0 and $\mathcal{J}(\cdot)$ tends to 1.
Equation~(\ref{eq:modifiedcost1}) can be minimised using a variety of planning techniques,  such as rapidly-exploring random trees (RRTs) \citep{bib:lavalle_1998}, probabilistic roadmaps (PRM) \citep{bib:kavraki_1996}, differential dynamic programming (DDP) \citep{bib:jacobson_book_1970} or sequential dynamic programming (SDP) \citep{bib:betts_book_2001}. Next, we show the implementation of our information gathering algorithm by using a modified version of a PRM to cope with high DoF manipulators and non-convex objects.

\begin{figure}[!t]
\centerline{
\includegraphics[width=.95\columnwidth]{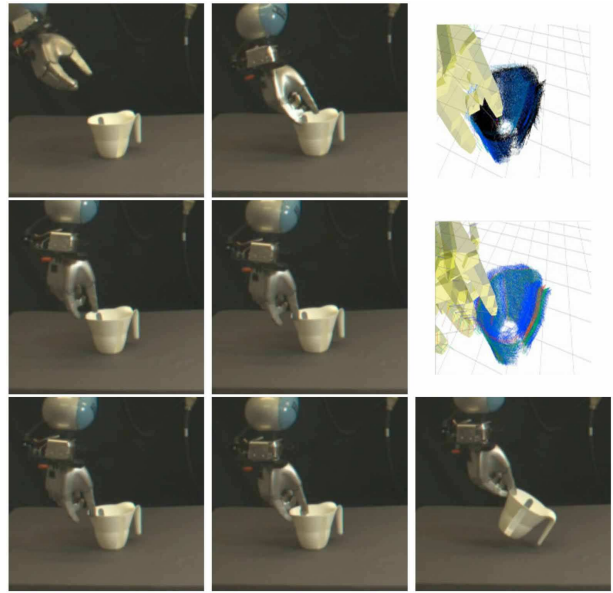}
}

\caption[Information gain on Boris]{The sequence of images presents an interesting case in which the initial belief state does not cover the ground truth. IR3ne however is capable of converging to a grasp in three iterations. The simulated images show the belief update as a PF. The colour of each hypothesis is associated with its likelihood, from red (high likelihood) to black (zero likelihood). Top row: due to an erroneous localisation the first grasp attempt collides with the plastic jun on the rim (middle). The contact is used to refine the belief state but, since none of the hypotheses matches the contact, the hypotheses have all low probability associated (right image). Middle row: a second attempt fails but this time the contact allows Boris to localise the object and, finally, to grasp it (third row).}
\label{fig:irene_boris_1}
\end{figure}

\section{Information-Reward based Re-Planning Algorithm}
\label{sec:implementation}

This section provides a detailed explanation of the principal components in our architecture.
First, Sec.~\ref{sec:ch06:observational_model} presents our observational model for tactile observations. 
Sections \ref{sec:ch06:heuristic} and \ref{sec:motion_planning} show our modified PRM algorithm which allows us to efficiently plan reach-to-grasp trajectories for high DoF manipulators, as well as compute the information reward function, described in Eq.~(\ref{eq:modifiedcost1}), for each node of the PRM. Section~\ref{sec:non_convex} describes our PCOM-based collision detection algorithm. Sections~\ref{sec:belief_update} and \ref{sec:replanning} show how unexpected contacts are integrated into the belief state and the re-planning phase. Finally, Sec.~\ref{sec:terminal} presents the terminal conditions for the algorithm.

\subsection{An observation model for tactile contacts}\label{sec:ch06:observational_model}

The manipulator is composed of a set of rigid links organised in a kinematic chain and tree, comprising an {\em arm}, $L_0$, and a set of $J$ multi-joint {\em fingers}, denoted $\{L_j\}_{j=1}^J$. 
We are only interested in the observations made by the fingers. Thus, we define $\mathbf{x}_{L_j}$, or simply $\mathbf{x}_j$, as the joint configuration of the $j^{th}$ finger. Through forward kinematics we know the relative pose of the fingertip, denoted $\mathcal{W}(\mathbf{x}_j)\in SE(3)$.

The expected observation (contact) is related to the probability of a physical contact between the finger links and the object. We assume that this probability decays exponentially, based on the distance, $d_{ji}$, between each finger and the closest surface of the object assumed to be in pose $p_i$ (the computation of $d_{ji}$ is explain in Sec.\ref{sec:non_convex}). The observation model is limited to contacts which may occur on the internal surface of fingers. This directly affects the planner, which rewards trajectories that would generate contacts on the fingertips rather than on the back side of the fingers. Therefore we rewrite Eq.(\ref{eq:h}) as 
$\mathbf{z}_i=g(\mathbf{x}, p_i)=\prod_{j}{\varphi(\mathbf{x}_j, p_i)}$, where
\begin{equation}\label{eq:obs_model}
\varphi(\mathbf{x}_j, p_i)=
\begin{cases}
  \eta e^{(-\lambda d_{ji})} & \text{if }d_{ji}\leq d_{max} \\
	& \text{and }\langle \mathbf{n}_{x_j},\mathbf{\hat{n}}_{p_i}\rangle < 0 \\
  0 & \text{otherwise}
\end{cases}
\end{equation}
where $\langle \mathbf{n}_{x_j},\mathbf{\hat{n}}_{p_i}\rangle$ is the inner product of, respectively, the $j^{th}$ fingertip's normal and the estimated object surface's normal, and $d_{max}$ describes a maximum range in which the probability of reading a contact is non-zero, $\eta$ is a normaliser. 


\begin{figure}[!t]
\centerline{
\includegraphics[width=.95\columnwidth]{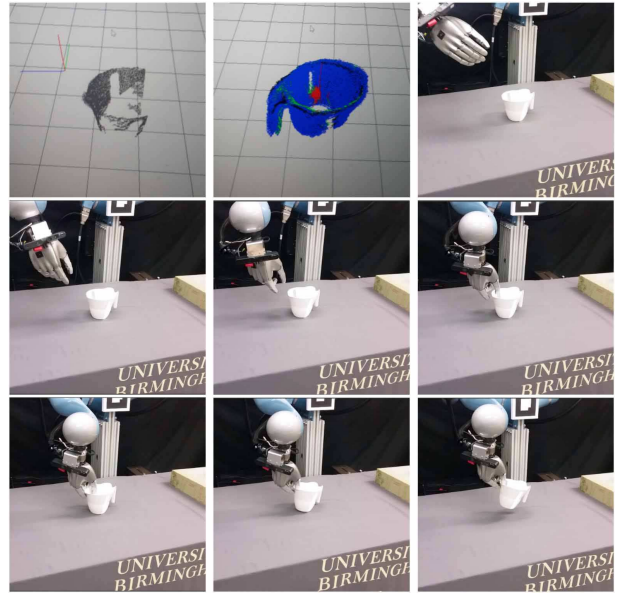}
}

\caption[Information gain on Boris]{Example of plan execution for IR3ne. The top row shows: the partial query point cloud (three merged views),  the belief state (as sub-sampled hypotheses (blue), mean pose (green) and ground truth (black)), the real pose of the object. This example shows the worst case in which the query point cloud covers only a 22.5\% of the model and the handle is not visible. Note that the ground truth (black point cloud, middle image) is also estimated with the wrong orientation. Nevertheless, IR3NE executes the planned trajectory (middle row) and achieves a grasp. In the bottom-right image, Boris lifts the jug successfully.}
\label{fig:irene_boris_2}
\end{figure} 

\subsection{Implementation of our planner IR3ne}\label{sec:ch06:heuristic}

The approach here incorporates the expected (tactile) value of information into the metric for the underlying physical space. This results in warping distances so as to create a non-Euclidean space, in which trajectories that maximise information gain are less costly\footnote{In this sense, the cost function we propose works similarly to the Mahalanobis distance.}. 

We employ a modified version of Probabilistic Roadmap (PRM) \citep{bib:kavraki_1996} to sample the reachable configuration space of the robot. The original PRM algorithm is composed of two phases: i) a \emph{learning phase}, in which a connected graph $G$ of obstacle-free configurations of the robot is generated and, ii) a \emph{query phase}, which finds a path between a given pair of configurations $\mathbf{x}_{root},\mathbf{x}_{goal}$. However the computational cost for the learning phase grows fast with respect to the dimensionality of the space. Because of this we incrementally build connections between neighbouring nodes during the query phase. 

Given a pair $\langle \mathbf{x}_{root}, w_{goal}\rangle$, which describes the root state in configuration space, $\mathbf{x}_{root}\in\mathbb{R}^n$, and the goal state in workspace, $w_{goal}\in SE(3)$, of the trajectory, we use an A* algorithm to find a minimum cost trajectory on the graph $G$. 
The A* algorithm selects the path which minimises the objective function
\begin{equation}\label{eq:cost}
c(\mathbf{x}) = c_1(\mathbf{x},\mathbf{x}_{root}) + c_2(\mathbf{x},\mathbf{x}',\hat{\mathbf{x}}_{goal})
\end{equation} 
where $\mathbf{x},\mathbf{x}'\in\mathbb{R}^n$ and $\mathbf{x}'\in Neighbour(\mathbf{x})$, $\hat{\mathbf{x}}_{goal}$ is a reachable goal configuration for the robot computed by inverse kinematics. 
The function $c_1(\mathbf{x},\mathbf{x}_{root})$ is a cumulative travel distance to $\mathbf{x}$ from $\mathbf{x}_{root}$, while $c_2(\mathbf{x},\mathbf{x}',\hat{\mathbf{x}}_{goal})$ is a linear combination of the cost-to-go from $\mathbf{x}$ to a neighbour node $\mathbf{x}'$ and the expected cost-to-go from $\mathbf{x}'$ to the target, defined as:
\begin{equation}\label{eq:cost2}
c_2(\mathbf{x},\mathbf{x}',\hat{\mathbf{x}}_{goal})=\alpha d_{bound}(\mathbf{x},\mathbf{x}')+\beta d(\mathbf{x}',\hat{\mathbf{x}}_{goal})
\end{equation} 
where $\alpha,\beta\in [0,1]$, $d(\cdot)$ is a distance function in $SE(3)$, and $d_{bound}(\cdot)$ behaves as $d(\cdot)$ for neighbouring nodes, but returns $+\infty$ otherwise. Equation~(\ref{eq:cost}) does not violate the consistency and admissibility of the cost function for A* in Euclidean spaces. The algorithm BSP (Sec.~\ref{sec:experiments}) uses the above cost function to plan reach-to-grasp trajectories.

In IR3ne, we modify the heuristic $c_2(\cdot)$ in order to reward informative tactile explorations while attempting to reach the goal state (described as a target configuration of the manipulator), by embedding the information value as follows: 
\begin{equation}\label{eq:newcost2}
\begin{aligned}
\bar{c}_2(\mathbf{x},\mathbf{x}',\hat{\mathbf{x}}_{goal},A,p_{1:k})=&\alpha \mathcal{J}(\mathbf{x},\mathbf{x}',p_{1:k})d_{bound}(\mathbf{x},\mathbf{x}')\\
&+\beta d_A(\mathbf{x}',\hat{\mathbf{x}}_{goal})
\end{aligned}
\end{equation}
where $A$ is the diagonal covariance matrix of the sampled states, for any column vector $\mathbf{a},\mathbf{\mu}\in\mathbb{R}^n$, $d_A(\mathbf{a}, \mathbf{\mu})=\sqrt{(\mathbf{a}-\mathbf{\mu})^TA^{-1}(\mathbf{a}-\mathbf{\mu})}$ is the Mahalanobis distance centred in $\mu$ and $\mathcal{J}(\mathbf{x},\mathbf{x}',p_{1:k})\in(0,1]$ is a factor which rewards trajectories with a large difference between expected observations if the object is at the expected location, $p_1$, versus observations that would be expected if the object is at other poses, $p_{2:k}$, sampled from the distribution of poses associated with the object's positional uncertainty:
\begin{equation}\label{eq:modifiedcost}
\mathcal{J}(\mathbf{x},\mathbf{x}',p_{1:k})=\frac{1}{k-1}\sum_{i=2}^k{e^{-\Phi(\mathbf{x},\mathbf{x}',p_i)}}
\end{equation}
where:
\begin{equation}
\Phi(\mathbf{x},\mathbf{x}',p_i)=||g_t(\mathbf{x},\mathbf{x}',p_i)-g_t(\mathbf{x},\mathbf{x}',p_1)||_2
\end{equation}
for each $i\in[2,k]$, and $g_t(\mathbf{x},\mathbf{x}',p_i)$ is sequence of probabilities of reading a contact travelling from state $\mathbf{x}$ to $\mathbf{x}'$. In this implementation $g_t(\mathbf{x},\mathbf{x}',p_i)=g(\mathbf{x}',p_i)$. In other words, we evaluate the probability of making a contact while moving from state $\mathbf{x}$ to $\mathbf{x}'$ as the probability of making a contact only in the next state $\mathbf{x}'$. Figure~\ref{fig:obs_model} shows the effects of the reward function $\mathcal{J}(\cdot)$ in a 2D example.

Note that this observation model is designed to conserve Eq.(\ref{eq:newcost2}) as in Eq.(\ref{eq:cost2}) when the probability of observing a tactile contact is zero. In fact, for robot configurations in which the distance to the sampled poses is larger than a threshold, $d_{max}$, the cost function $\mathcal{J}(\cdot)$  is equal to 1. However we also encode uncertainty in $d_A(\cdot)$, which evaluates the expected distance to the goal configuration. In this way the planner also copes with pose uncertainty at the early stages of the trajectory, when the robot is still too far away from the object to observe any contacts. 

\subsection{Planning for dexterous manipulator}\label{sec:motion_planning}

Several hierarchical path planning techniques have been developed over the past years to enable high DoF robots to move in complex environments. Typically, the hierarchy makes use of: i) a global planner and ii) a local planner. The global planner behaves as a sample-based approach to probe a grid-based environment to quickly find a coarse solution. Subsequently, the local planner behaves as an optimisation procedure to refine the solution. This approach has been exploited in numerous variations, but mostly focussed on mobile robots, as in~\citep{bib:zhu_1991}. For dexterous grasping, however, the path planning problem is usually addressed by constraining the plan on a set of manifolds that capture the human hand's workspace, as in~\citep{bib:rosell_2011}.  

In contrast, we adopt a hierarchical path planner that enables us to plan dexterous reach-to-grasp trajectories up to 21 DoF. First a PRM is constructed only in the arm configuration space in order to find a global path between the $\mathbf{x}_{root},\hat{\mathbf{x}}_{goal}$. It is worth noticing that, in this phase, the rest of the joints of the manipulator are interpolated in order to have a smooth passage from $\mathbf{x}_{root}$ to $\hat{\mathbf{x}}_{goal}$. Then the planned trajectory is refined by constructing a new local PRM. The local PRM only samples configurations nearby the waypoints of the global trajectory.  However the local PRM has a higher dimensional configuration space to represent the whole of the manipulator (e.g. arm + hand joint space).
Subsequently we employ a Differential Evolution (DE) procedure to optimise the trajectory and generate a smoother transition from one configuration to the next.

This approach allows us to integrate the information gain cost function described in Eq.(~\ref{eq:newcost2}) into both levels of the hierarchy. Hence, we enable the robot to adjust the trajectory of each of its fingers if more information is expected to be gained.

\subsection{Planning a dexterous grasping trajectory for non-convex objects}\label{sec:non_convex}

\begin{figure}[!t]
\centerline{
\includegraphics[width=.59\columnwidth]{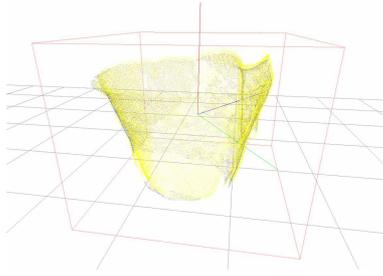}
}
\caption[Hierarchical collision detection]{The image shows the target object to be grasped. In this case the object is a jug. The grey point cloud represents the ground truth pose of the object, as it was acquired by a noiseless input source. The yellow point cloud identifies the best pose estimate. The yellow point cloud is organised as a KD-Tree for faster collision detections. The red box represents the bounding box which avoids unnecessary collision checking between the object and the robot's links when they are far apart.}
\label{fig:bounding_box}
\end{figure}

The problem of efficiently computing intersection-depth between two intersecting objects is still an open problem especially in the general case, in which at least one of the two object is a non-convex polyhedron. Our approach is based on a \emph{penetration depth} (PD) algorithm, which is a measure of the overlap between two objects. The PD algorithm for non-convex polyhedra has a lower bound for the worst-case time complexity of $O((m+n)\log^2(m+n))$, where $m$ and $n$ are the number of faces that compose the objects~\citep{bib:weller_2013}.

In contrast, we represent the robot's bodies by an open-chain of convex polyhedra, and the object to be grasped as a 2-level structure (Fig.~\ref{fig:bounding_box}). The first level of this structure wraps the PCOM in a bounding box. This level can efficiently avoid checking collisions when the robot's links are far from the object to be grasped. The second level contains the PCOM organised in a KD-Tree~\citep{bib:muja_2014}. 

Our approach works as follows. Each link $L_j$ has a pre-computed boundary $B_j$ defined as a set of triangle meshes and a reference frame $O_j=\{\mathbf{p}_j,q_j|\mathbf{p}_j\in\mathbb{R}^3,q_j\in SO(3)\}$ placed at the centre of mass of the polyhedron. Each triangle mesh, $t_i$, is composed of three vertices $\mathbf{v}_{i1}, \mathbf{v}_{i2}, \mathbf{v}_{i3}\in\mathbb{R}^3$ and a normal $\mathbf{n}_{i}\in\mathbb{R}^3$, such that $\mathbf{n}_i=\frac{(\mathbf{v}_{i2}-\mathbf{v}_{i1})(\mathbf{v}_{i3}-\mathbf{v}_{i1})}{||(\mathbf{v}_{i2}-\mathbf{v}_{i1})(\mathbf{v}_{i3}-\mathbf{v}_{i1})||}$ is chosen with the direction that points outside the polyhedron. Given a set of nearest points, $A$, to the link pose $\mathbf{p}_j$, we compute the penetration depth value for each triangle, $t_i$, as the average depth 
$$
d_{t_i}= \frac{1}{|A|}{\sum_{\mathbf{a}\in A}{||\mathbf{v}_{i1}||_{\mathbf{p}_j}-\mathbf{n}_i^\top \mathbf{a}}}
$$
where $|A|$ is the cardinality of the set $A$, $||\mathbf{v}_{i1}||_{\mathbf{p}_j}$ is the distance of $t_i$ from the reference frame, approximated as the distance to $t_i$'s first vertex, and $\mathbf{n}_i^\top \mathbf{a}$ is a dot product. 
The depth associated with the bound $B_j$ is computed as $d_{j}=\min_{t_i\in B_j}\{d_{t_i}\}$.

The value $d_{k}$ is not a distance, but it assumes negative values if the majority of points in $A$ lie outside the polyhedron defined by $B_k$.

Our planning process uses the PD algorithm in two cases: i) to reject PRM nodes in collision with an object and ii) to compute the information value for each node of the PRM. 
In the latter case, we rewrite $d_{ji}$ from Eq.(\ref{eq:obs_model}) as
\begin{equation}\label{eq:bound_distance}
  d_{ji}=
\begin{cases}
 |d_j| & \text{if }d_j<0 \\
  0 & \text{otherwise.}
\end{cases}
\end{equation}

The worst-case complexity of the overall PD algorithm is dominated by the estimation of the closest point in the PCOM, which is $O(|B|\log n)$, where $n$ is the number of points in the PCOM~\citep{bib:muja_2014} and $|B|$ the number of bounds in the hand. The same happens for the space complexity which is bounded by storing the PCOM in the kd-tree, $O(n)$.

\begin{figure}[!t]
\centering {
       \includegraphics[width=.95\columnwidth]{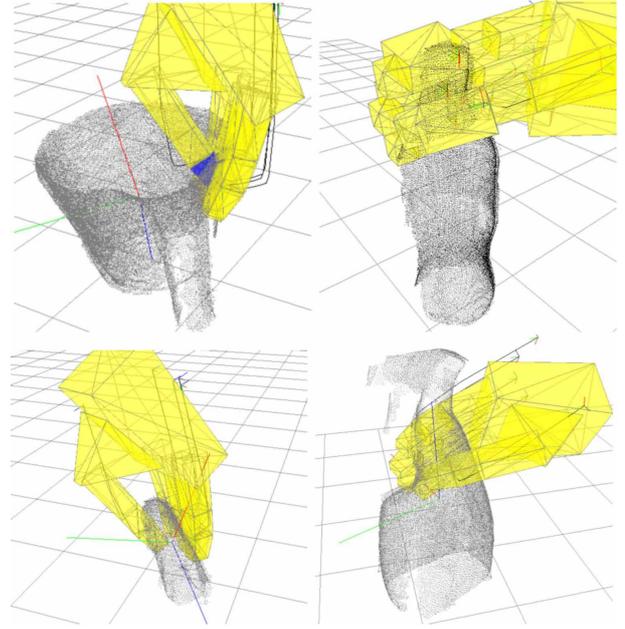}
 }
\caption[Object models with associated grasp]{Models of the objects and their associated grasps. The top left image shows a rim grasp on a jug. In this case, the target grasp requires the thumb to be placed on the internal surface of the object, thereby penetrating inside the convex hull. The top right image shows a top grasp on a bottle of coke. The bottom row images show respectively a rim grasp on a stapler and a Mr Muscle spray bottle. The grasp configurations are computed using the method described in~\citep{bib:kopicki_2014}.}
\label{fig:07:grasps}
\end{figure}

\begin{figure*}[!t]
\centerline{
\includegraphics[width=.95\textwidth]{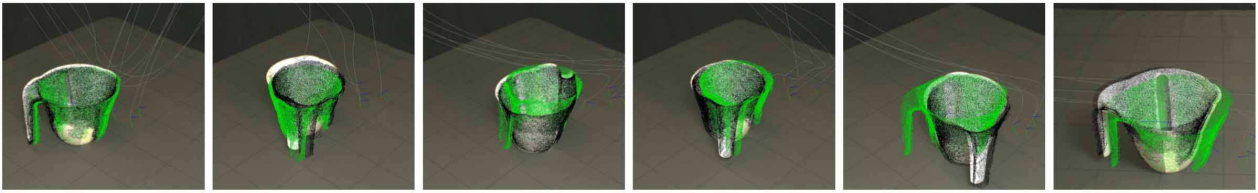}
}

\caption{Initial pose uncertainty. The sequence of images shows some examples of initial pose estimation during the experiments on Boris. The simulated point cloud overlays the real plastic jug to show the misalignment between the real pose of the object, the ground truth (black point cloud) used to evaluate the simulated results and the initial pose estimate (green point cloud), to which the algorithm attempt the grasp. The lines show the planned trajectory for each finger.}
\label{fig:07:jug_boris_trials}
\end{figure*}

\subsection{Belief update}\label{sec:belief_update}

When tactile observations occur the algorithm refines the current belief state. We think of the reach-to-grasp trajectory as composed of two parts: i) the approach trajectory which leads to a pre-grasp configuration of the robot in which the fingers generally cage the object to be grasped without generating any contact, and ii) a finger closing trajectory which moves the fingers into contact and generates a force closure grasp. In this way any contact which occurs during the approach trajectory is considered as an unexpected observation. Similarly an insufficient number of contacts for a force closure at the end of a grasping trajectory can also be used as an observation. In our implementation, the belief is updated assuming deterministic dynamics.

At any time step, $t$, a belief state $b^t$, supported by a set of particle $\mathcal{S}^t=\{y_j^t\}_{j=1}^{N}$, represents a density function over poses for the object to be grasped. Once a trajectory is executed and a real (unexpected) observation $z^t$ is detected, the belief state is updated by integrating the new acquired information. We denote $b^{t+1}(y_j^t)$ the updated belief, which is computed following the Bayes' rule 
\begin{equation}
b^{t+1}(y_j^t)=\eta P(z=z^t|x^t,y_j^t)b^t(y_j^t)
\end{equation}
in which $\eta$ is the normalising factor, and then re-sampling is performed to generate \emph{a posteriori} distribution $b^{t+1}$ as new set of particles $\mathcal{S}^{t+1}=\{y_j^{t+1}\}_{j=1}^{N}$.


\subsection{Re-planning}\label{sec:replanning}


We need to rely on sensory feedback during the execution of the planned trajectory in order to detect whether or not unexpected observations occur. This triggers a belief update, using the observation gathered at execution-time, and consequently a re-planning phase. 
In our experiments, the algorithm uses torque sensors based on the current draw at each joint of the robot's hand to detect whether or not a link of the hand is in contact with the environment.  

\subsection{Terminal conditions}\label{sec:terminal}

Our algorithm terminates its execution when no unexpected contacts occur and the target grasp is achieved. Since we do not have mesh models of the object to be grasped we cannot rely on force closure measures to signal successful termination of the algorithm. Nonetheless, in simulation it is possible to measure if the planned contacts between the fingers and the object's point cloud have been achieved. On the real robot, the success of the grasp is evaluated by lifting the object. If the robot can hold the object, the grasp is considered successful.

\section{Experiments}
\label{sec:experiments}

Three strategies for planning dexterous reach-to-grasp trajectories are evaluated in both a half-humanoid robotic platform, Boris, and a simulated environment:
\begin{itemize}
\item PRM: an open-loop motion planner that plans trajectories towards the expected object pose without re-planning.
\item BSP: a sequential belief state re-planner that tracks high dimensional beliefs. 
\item IR3ne: a sequential information-reward based re-planner that maximises the information gain by planning in a fixed dimensional hypothesis space. 
\end{itemize}

Table~\ref{tab:algorithms} summarises the differences between the algorithms, by showing which features are implemented. The features represent i) Motion Planning (MP), the ability of the algorithm to compute a free-collision trajectory for a robot manipulator, ii) Belief Planning (BP), the ability of updating the high-dimensional belief state by integrating new observations, iii) Hypothesis Based Planning (HBP), the ability of planning in a fixed dimensionality space underlying the belief state, iv) Simultaneous Information Gain \& Grasping (SIGG) and v) Re-Planning (RP).

Sec~\ref{sec:07:boris_results} presents empirical results collected using Boris (Fig.~\ref{fig:boris}), in which we tested the ability of each algorithm to pick up an object in a real scenario under the presence of pose uncertainty, due to incomplete perception abilities. The results are based on a total of 30 trials (10 per strategy) on a plastic jug. 

The experimental results show that:
\begin{itemize}
\item Sequential re-planning achieves higher grasp success rates than open-loop grasping from vision.
\item Trajectories that maximise information gain (IR3ne) generate contacts with the object that are more informative (to locate it).
\item More informative contacts lead to fewer re-planning iterations to converge to a grasp.
\item Trajectories that maximise information gain increase up to 50\% the proportion of grasps on a first attempt.
\item On the real robot, IR3ne increases grasp reliability over BSP.
\end{itemize}

\subsection{Experiments in a virtual environment}\label{sec:07:simulated_results}

The experiments are composed of 120 trials per object in a virtual environment, and they are run over four different objects: a jug, a coke bottle, a stapler and a Mr Muscle spray bottle. The algorithm has a model of the object to be grasped, in the form of a dense point cloud, acquired by scanning the object with a depth camera. Figure~\ref{fig:07:grasps} shows the PCOM of each simulated object, and the associated target grasp configurations. For these experiments, the dense models are composed of 7 single-view point clouds.

Four different conditions have been tested. In each condition we randomly selected either 1, 3, 5 or 7 views of the object, in order to simulate a real situation in which the robot's depth camera observes smaller or larger parts of the object. Hence, each trial has a different initial probability density over the object pose which depends on how much of the point cloud is visible. The density is computed by using our model-fitting algorithm as described in Sec.~\ref{sec:belief_state_estimation} with 1,000 surflet features sampled, which makes the estimation very fast but not accurate.

Figure~\ref{fig:07:sim_results} summarises the data collected in our experiments. All the results are compared with respect to the model coverage in percentage terms, i.e. one view typically covers 20\% of the object. A ground truth pose for the object is also calculated for each object by using our model-fitting algorithm (Sec.~\ref{sec:belief_state_estimation}), but sampling 500,000 surflet features. The algorithms have no knowledge of the ground truth pose, however in the virtual environment the ground truth is used to model the real object location, and is used to trigger simulated contacts with the robot hand. These simulated contacts cause the sequential re-planning algorithm to stop and update the belief state. 

\subsection{Discussion of simulated experiments}\label{sec:discussions_sim}

\begin{table}[t]
\centering
\caption{Feature comparison between the algorithms. We compare 3 methods: PRM, BSP, and IR3ne (see full description in Sec.~\ref{sec:experiments}). The features are: Motion Planning (MP), the ability of the algorithm to compute a free-collision trajectory for a robot manipulator; Belief Planning (BP), the ability of tracking the high-dimensional belief state by integrating new observations; Hypothesis Based Planning (HBP), the ability of planning in a fixed dimensionality space underlying the belief state; Simultaneous Information Gain \& Grasping (SIGG); and Re-Planning (RP).}
\label{tab:algorithms}
\begin{tabular}{ | l | c | c | c | c | c |}
\hline
Method & MP & BS & HBP & SIGG & RP \\ 
\hline
PRM & \checkmark & \xmark & \xmark & \xmark & \xmark \\ \hline
BSP & \checkmark & \checkmark & \xmark &\xmark & \checkmark \\ \hline
IR3ne & \checkmark & \checkmark & \checkmark & \checkmark & \checkmark \\ \hline
\end{tabular}
\end{table}

Section~\ref{sec:07:simulated_results} presents the results collected on 120 trials in simulation for four objects: a jug, a bottle of coke, a stapler and a Mr Muscle spray bottle. In these experiments, the aim is to evaluate the ability of the three strategies to localise the target object as well as converge to a target grasp configuration with artificially induced pose uncertainty. 

\begin{figure*}[t]
\centerline{
\subfloat[]{\includegraphics[width=.325 \textwidth]{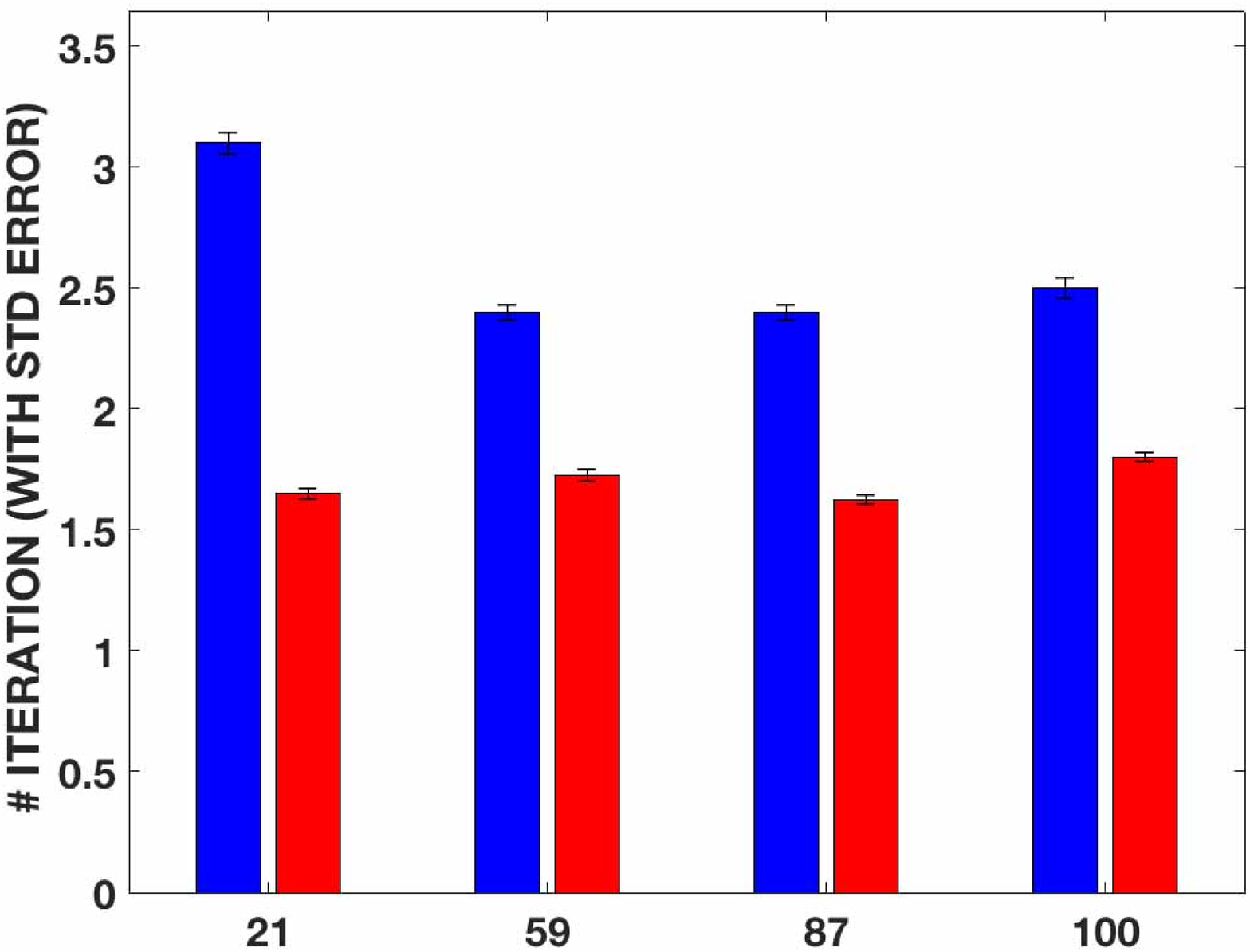}\label{subfig:sim_iterations}}
\subfloat[]{\includegraphics[width=.325 \textwidth]{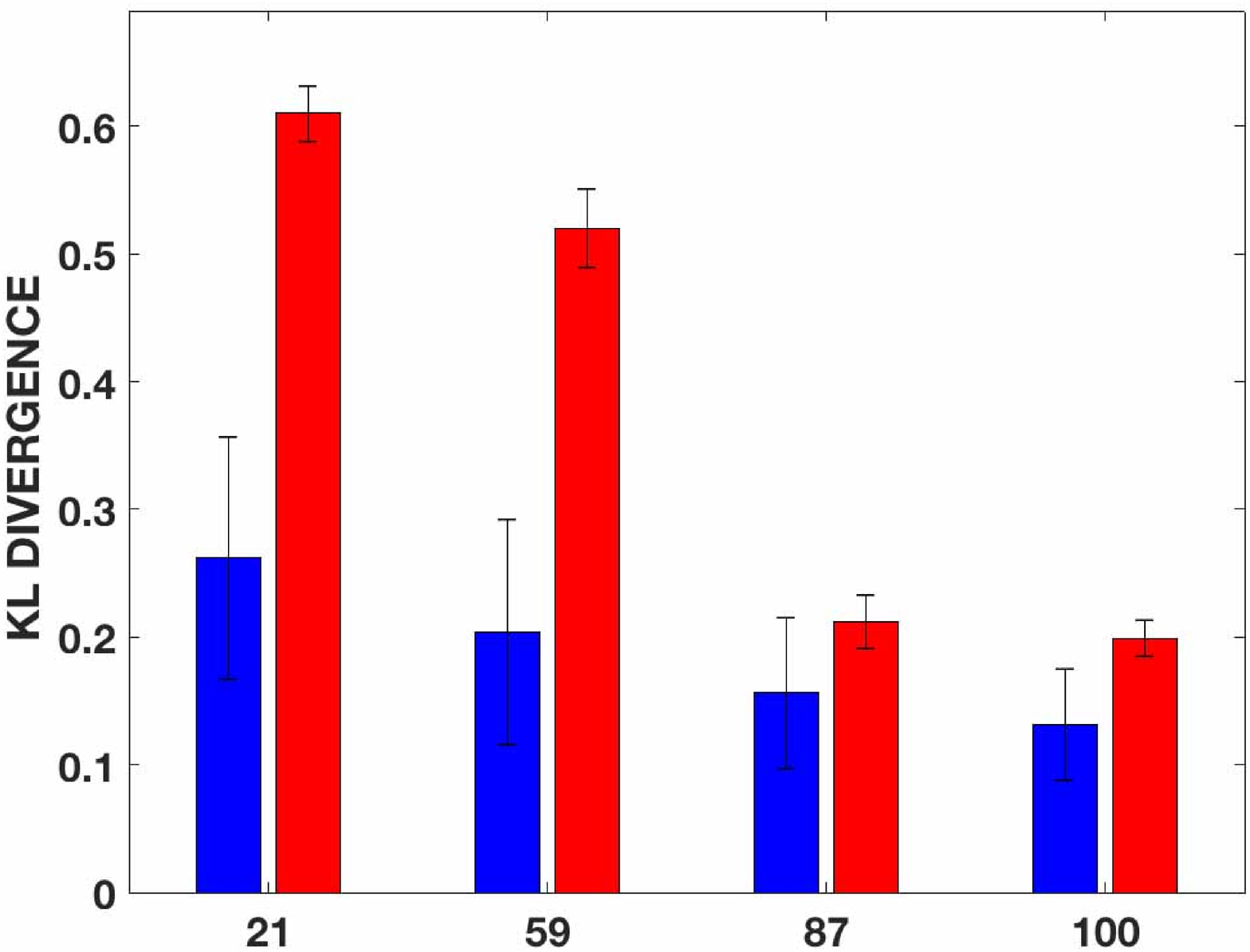}\label{subfig:sim_touch}}
\subfloat[]{\includegraphics[width=.325 \textwidth]{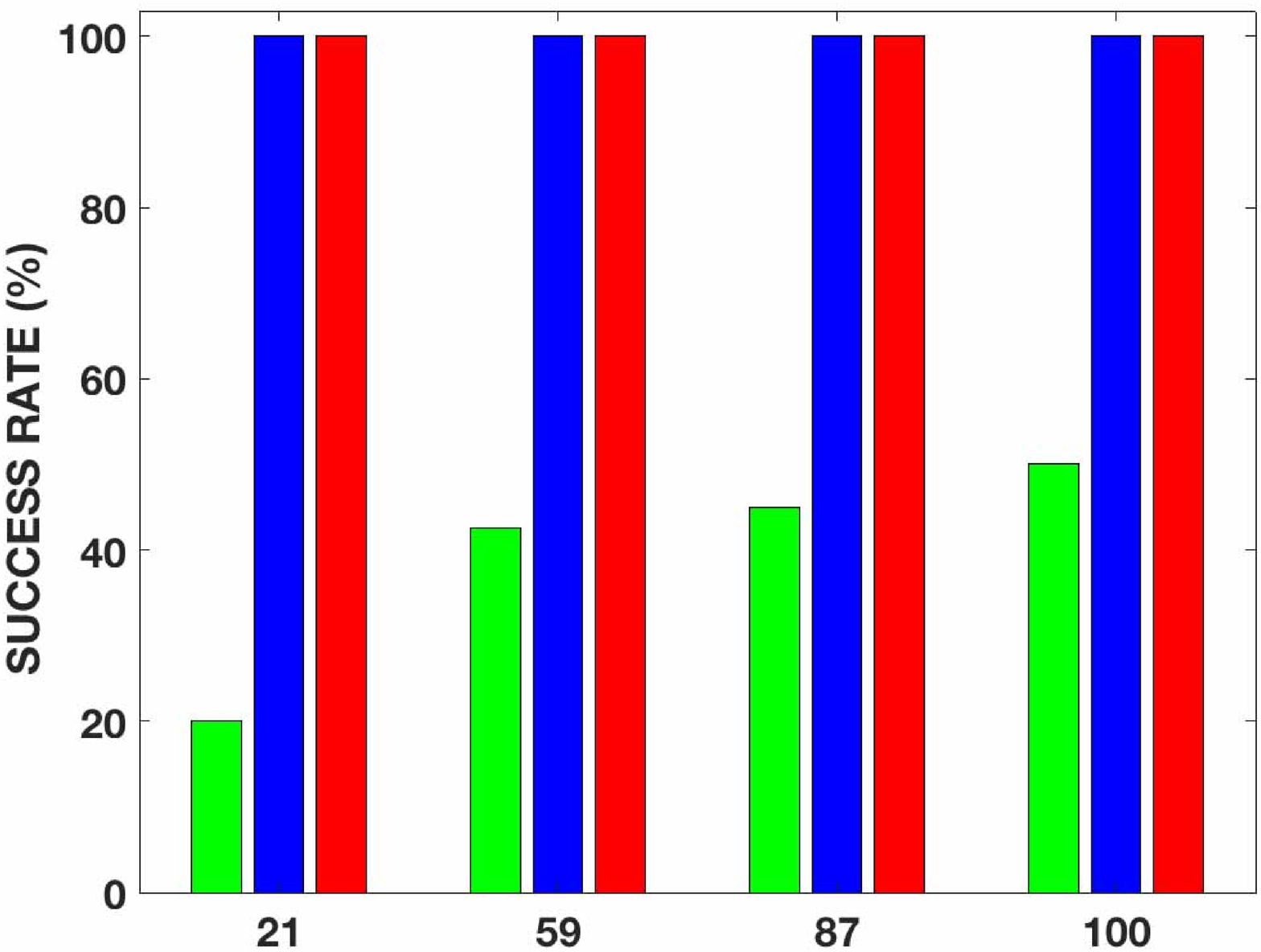}\label{subfig:sim_success}}
}
\caption[Simulated results]{Simulated results for 120 trials on 4 objects. Three strategies were tested: PRM (green), BSP (blue), IR3ne (red). All the results are plotted against the initial percentage of model coverage for a total of four conditions. Figure (a) shows the number of iterations,(b) the information gain as KL divergence (Eq.~\ref{eq:kl}), and (c) the success rate. See Sec~\ref{sec:discussions_sim} for an extended discussion.
}
\label{fig:07:sim_results}
\end{figure*}

The results in Fig.~\ref{fig:07:sim_results} confirm that i) integrating the unexpected contacts into a belief state before attempting a new reach-to-grasp trajectory increases the success rate of grasping (Fig.~\ref{subfig:sim_iterations}) and ii) that information rewarded trajectories requires less re-planning iterations than conventional planning methods (Fig.~\ref{subfig:sim_success}). 

In contrast to typical belief space planning algorithms that separate exploratory and grasping actions and plans how to sequence them, our experimental results show that we do need to feed to the algorithm a confidence threshold to ensure a global pose estimation accurate enough to attempt a grasp. 

Figure~\ref{subfig:sim_touch} presents the expected information gain at each contact. We compute the information gain as follows
\begin{equation}\label{eq:kl}
D_{KL}(b^{t+1}||b^t)=-\sum_{i=1}^{k}{b^{t+1}(p^{t+1}_i)\log{\frac{b^t(p^t_i)}{b^{t+1}(p^{t+1}_i)}}}
\end{equation}
in which $D_{KL}$ is the Kullback-Leibler (KL) divergence between the pre- and post- contact belief states. Each particle $p_i$ is an element of the respective (i.e. pre- or post-) set of hypotheses as described in Eq.~\ref{eq:hypotheses}, and $b(p_i)$ is the likelihood associated with $p_i$ by the density $b$. A KL divergence of $0$ indicates that we can expect similar, if not the same, behaviour of two different distributions, while a KL divergence of $1$ indicates that the two distributions behave in such a different manner that the expectation given the first distribution approaches zero.
The results confirm that the contacts produced by IR3ne are significantly more informative than the ones generated by a conventional planner, especially when the uncertainty is wider. Also, the much smaller standard deviations for IR3ne compared to BSP in Fig.~\ref{subfig:sim_touch} show a more consistent and systematic acquisition of information from IR3ne throughout the trials. 

\subsection{Experiments on the robot Boris}\label{sec:07:boris_results}

We empirically demonstrated the ability of Boris of picking up objects in the presence of pose uncertainty. The pose uncertainty encodes the perceptual handicap of the robot, which leads to erroneous pose estimations when the object is poorly visible.

The results for the PRM strategy are extrapolated from BSP, since these two approaches construct a reach-to-grasp trajectory minimising the same cost function, therefore if BSP achieves (or fails to achieve) a grasp at the first iteration, the PRM also would succeed (or fail) equally. 

The data has been collected with the following procedure. The PCOM is acquired by scanning the operational workspace with a depth camera from 6 different views. These views are aligned, registered to generate a single PCOM.
In each trial, the target object is placed in a different configuration on a table in front of the robot, and a point cloud is acquired by scanning the operational workspace from 3 different views. Once these views are pre-processed, a belief state with a set of 5 particles is estimated as described in Sec.~\ref{sec:belief_state_estimation}. 

From the visible point cloud of the object, we compute the target grasp by adapting a pinch with support grasp learned on a bowl using the method described in~\citep{bib:kopicki_2015}. The training and test grasps are shown in Fig.~\ref{fig:07:bowllarge_pinchsupp}. Then a reach-to-grasp trajectory is computed and performed. A trial is considered successful if Boris converges to the target grasp configuration and lift the object from the table surface. 

Figure~\ref{fig:07:jug_boris_results} summarises the empirical results collected. 

\begin{figure*}[t]
\centerline{
\subfloat[]{\includegraphics[width=.325 \textwidth]{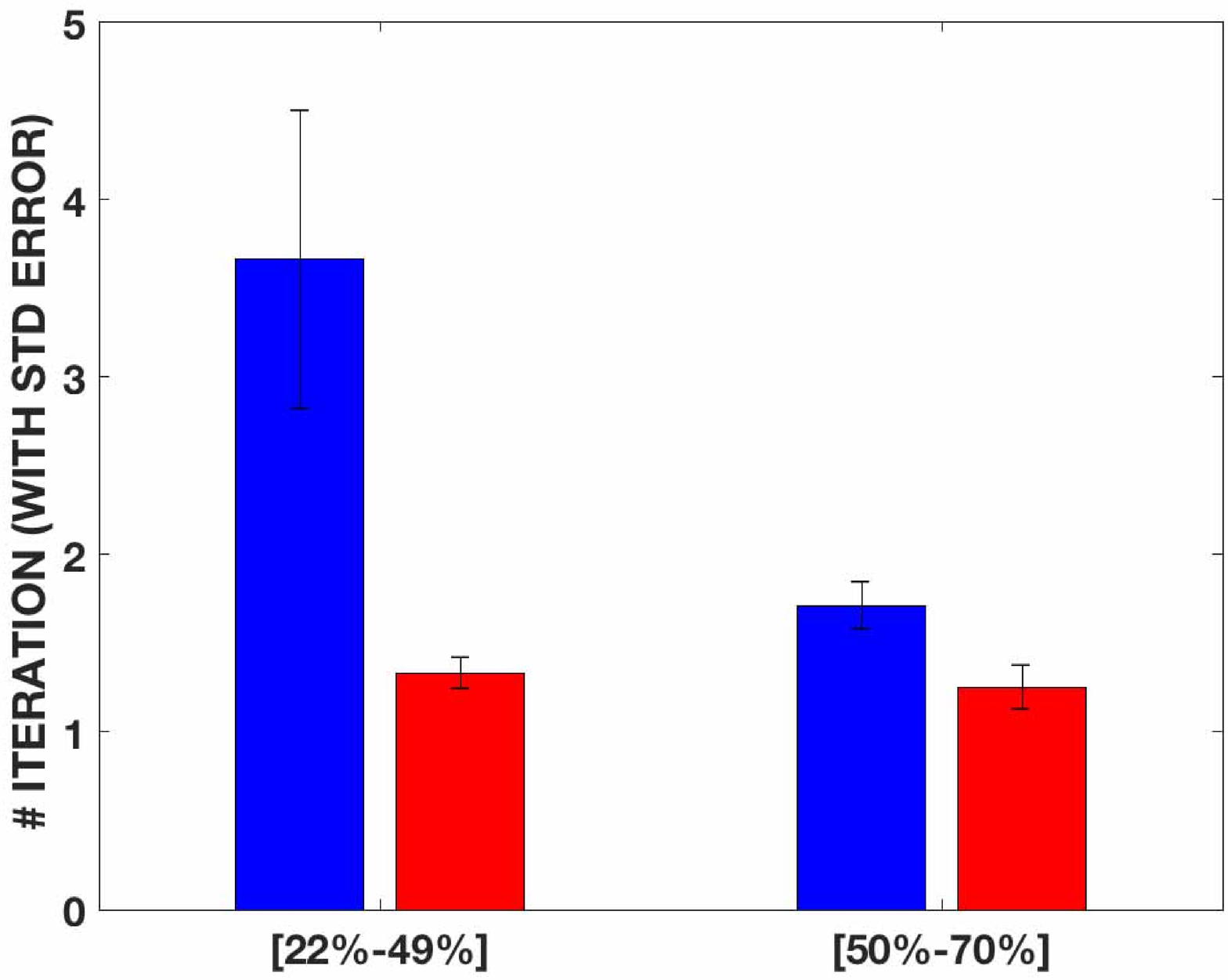}\label{subfig:real_iterations}}
\subfloat[]{\includegraphics[width=.325 \textwidth]{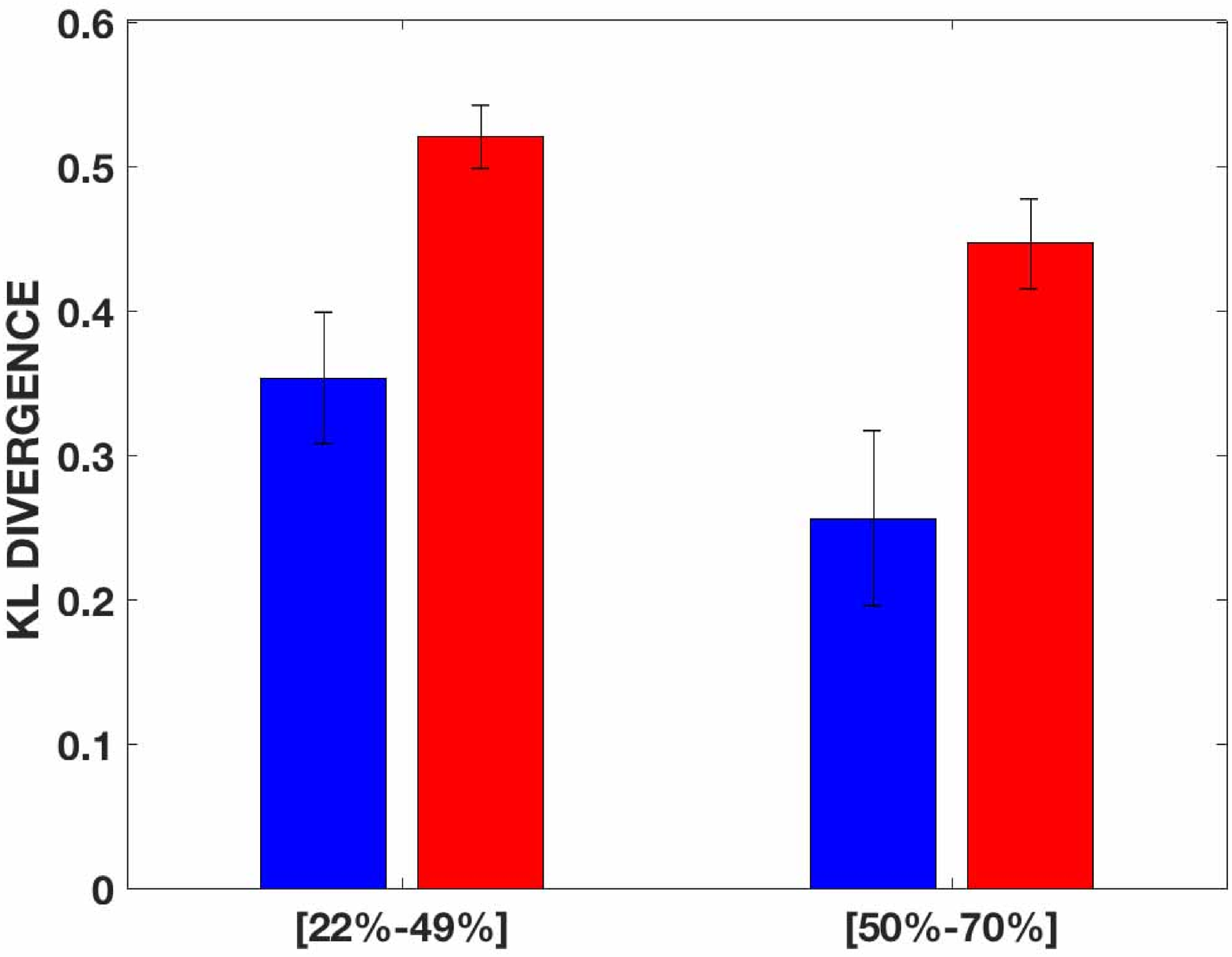}\label{subfig:real_touch}}
\subfloat[]{\includegraphics[width=.325 \textwidth]{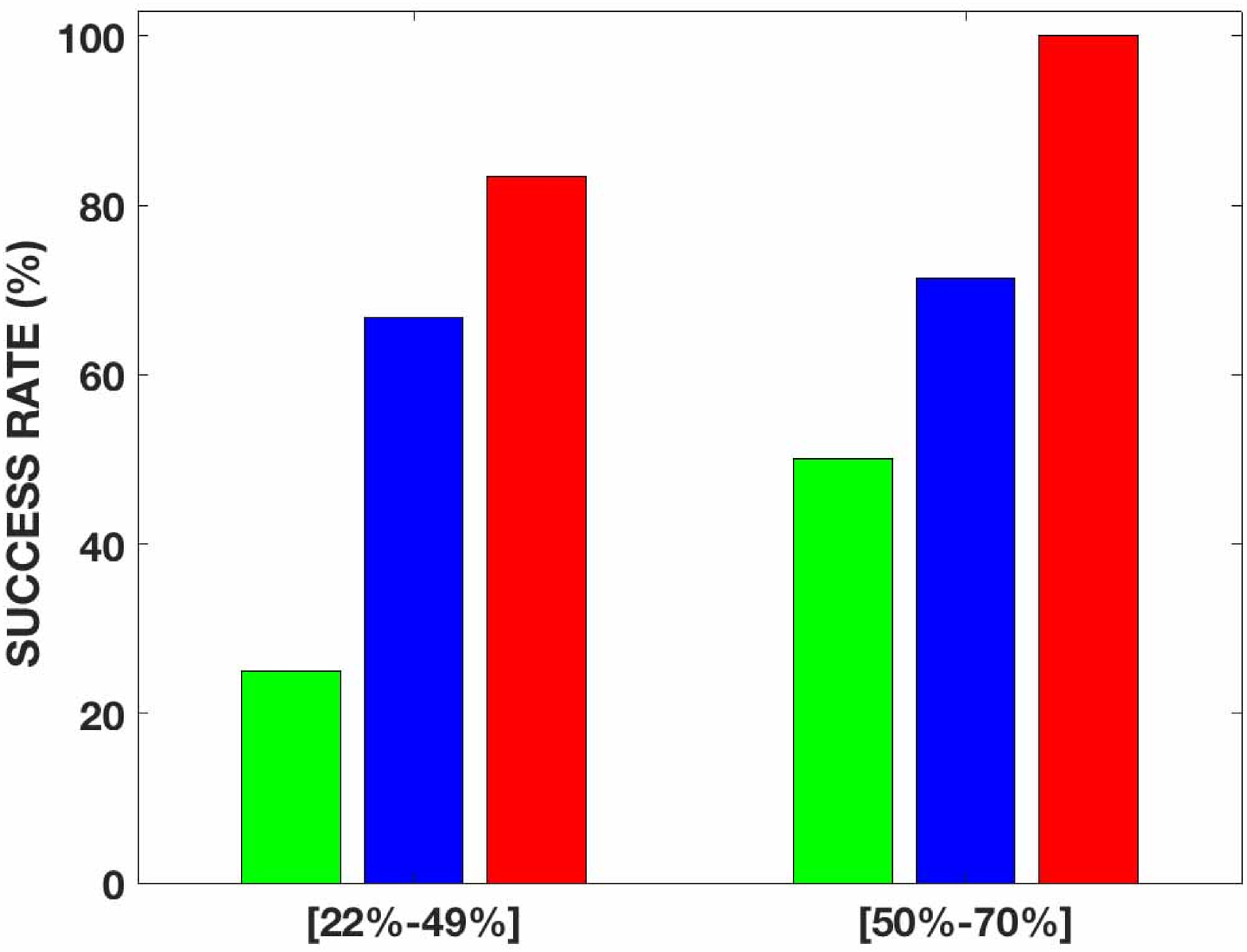}\label{subfig:real_success}}
}
\caption[Empirical results on Boris]{Empirical results on the Boris robot platform. The results refer to 30 trials (10 per strategy) on a plastic jug. Three strategies were used: PRM (green), BSP (blue), IR3ne (red). Figure (a) shows the number of iterations, (b) the information gain as KL divergence (Eq.~\ref{eq:kl}), and (c) the success rate. See Sec~\ref{sec:discussions_real} for an extended discussion.
}
\label{fig:07:jug_boris_results}
\end{figure*}

\begin{figure}[!t]
\centerline{
\includegraphics[width=.97\columnwidth]{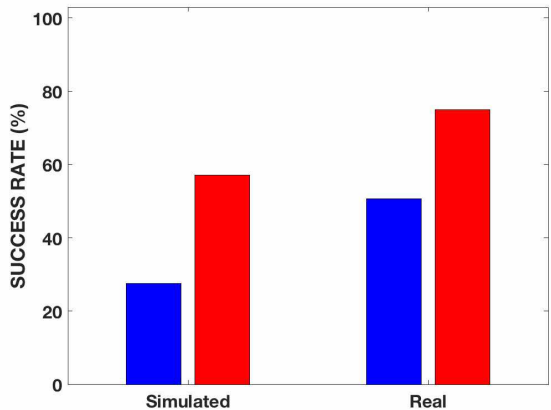}
}
\caption[First grasp attempt]{Successful rate of first grasp attempt.}
\label{fig:grasps}
\end{figure}

\subsection{Discussion of real experiments}\label{sec:discussions_real}

The empirical results in Fig.~\ref{fig:07:jug_boris_results} confirm the simulated data. Both (re-)planning algorithms are more efficiently converging to a grasp in the presence of not trivial pose uncertainty, i.e. errors in the pose estimation will not allow grasping at the first iteration with a conventional planner. Trajectories that maximise information gain are capable of generating informative contacts which outperform the information obtained by the other planner. This leads to fewer iterations and a better success rate.

Interestingly, IR3ne produces trajectories which are more robust with respect to the pose uncertainty, similar to the results presented in~\citep{bib:christopoulos}, by converging to the target grasp in a single attempt. Figure~\ref{fig:grasps} shows the IR3ne converges to a grasp in a single attempt up to 50\% more than a planner that does not explicitly reason on the belief state.



Figure~\ref{fig:irene_boris_2} is an interesting case in which the handle of the jug is not visible from the query point cloud and the object looks almost symmetric, which yields the mean-shift algorithm to a wrong pose estimation of about 180 degrees in the orientation. Nevertheless, IR3ne grasped in a single iteration. 

These empirical results have shown the validity of the sequential re-planning approach on a real scenario, however there are several issues that have to be addressed. First, the need for a predictive model for predicting how the target object moves after a contact occurs. Although the robot reliably stops after a contact is made, light objects, such as the plastic jug used in these trials, may be perturbed by the contact. For example, often we observe that the jug is tilted by a finger. The belief update by itself cannot deal with these cases and the resulting mean estimations are usually incorrect. However, the choice of point cloud models negatively affects the ability to develop predictive models.
Second, the lack of tactile sensors does not allow us to have a good estimate of which link of the finger experienced the contact. This results in an additional uncertainty in the belief update. 

Future work aims to address such issues by extending the sequential re-planning approaches to the use of tactile sensors and simple heuristics for an object-independent predictive model. The latter should depend on simple geometric properties of non-penetration between the object and the robot's fingers, so as to constrain the belief update and the mean pose estimate.

\section{Conclusion}
\label{sec:conclusion}

This paper has presented an approach for active tactile grasping that covers all stages: object modelling, state estimation from both vision and touch, grasp planning, re-planning, and execution. We have presented and tested a novel algorithm for planning reach-to-grasp trajectories under object pose uncertainty that actively rewards (tactile) information gathering. 

The main contribution of this work is to exploit active information gathering to the problem of dexterous robot grasping under non-Gaussian 6D uncertainty. We do so by embedding expected information value directly in the physical space. This results in an efficient planning algorithm that simultaneously gathers information while grasping.
We have also shown that our approach i) does not require a mesh model of the object to be grasped and ii) is not limited to grasp convex objects. We used a point cloud object model (PCOM) to represent the objects and presented i) a tactile observation model for PCOMs and ii) a point-cloud based collision detection for non-convex objects that can be efficiently integrated into the planner.


Experiments have shown that our approach (IR3ne) is effective in modifying reach-to-grasp trajectories to maximise the expected information gain while bringing the hand to a grasp configuration which is likely to succeed. In line with results in literature, we confirm that sequential re-planning achieves success rates which are higher than the success rates achieved with simple grasp attempts. Furthermore, IR3ne proves to require fewer re-planning iterations than conventional planning methods and improve up to 50\% the proportion of grasps that succeed on a first attempt. In contrast to conventional belief state planning algorithms, the experimental results confirm that our approach does not need to localise the object above a threshold of confidence to converge to a grasp.

The choice of a PCOM limits predictions of the object motion consequent to a contact. Even a light contact might move the object to a potentially unstable configuration. Future work will relax the assumption that tactile contacts have no physical effects. In addition, the incompleteness of the object model might have an effect on the detection of collisions during planning, such as generating trajectories which pass through the real target object. The current implementation of the algorithms interprets such contacts as unexpected observations which trigger re-planning. As future work, we have the intent to extend the planner to account for the likelihood that the unexpected contact was due to erroneous or incomplete shape information. Ideally, this would result in a SLAM problem, i.e., simultaneously localise an object while also mapping its shape, during iterative reach-to-grasp trajectories.   

\bibliographystyle{spmpscinat}
\bibliography{grasping_under_uncertainty}   
%

\end{document}